\documentclass{article} 
\usepackage{iclr2026_conference,times}

\usepackage{amsmath,amsfonts,bm}

\def\eqref#1{equation~\ref{#1}}

\def\1{\bm{1}}

\DeclareMathAlphabet{\mathsfit}{\encodingdefault}{\sfdefault}{m}{sl}
\SetMathAlphabet{\mathsfit}{bold}{\encodingdefault}{\sfdefault}{bx}{n}

\usepackage[T1]{fontenc}        
\usepackage[utf8]{inputenc}     
\usepackage{inconsolata}        
\usepackage{microtype}          

\usepackage{amsmath, amssymb, amsthm}
\usepackage{amsfonts}           
\usepackage{latexsym}
\usepackage{bbm}
\usepackage{pifont}             

\usepackage{booktabs}           
\usepackage{multirow}
\usepackage{colortbl}
\usepackage{tabularx}
\usepackage{makecell}
\usepackage{longtable}
\usepackage{siunitx}
\usepackage{diagbox}
\usepackage{graphicx}           
\usepackage{subcaption}
\usepackage{adjustbox}
\usepackage{wrapfig}

\usepackage[dvipsnames]{xcolor}
\definecolor{PineGreen}{RGB}{0,128,0}
\definecolor{OrangeRed}{RGB}{255,69,0}
\definecolor{mycolor}{RGB}{229,245,249}  
\definecolor{authorcolor}{RGB}{170, 68, 58}

\usepackage{algorithm}
\usepackage{algpseudocode}      
\usepackage{tgcursor}

\usepackage{enumitem}           
\usepackage[normalem]{ulem}     
\usepackage[most]{tcolorbox}    
\usepackage{fontawesome}
\usepackage{marvosym}
\usepackage{hyperref}
\usepackage{url}

\newtcolorbox{promptbox}{
    colback=teal!5!white,        
    coltitle=black,              
    fonttitle=\bfseries,         
    enhanced,                    
    borderline west={2pt}{0pt}{teal!75!black}, 
    frame hidden,                
    attach boxed title to top left={
        yshift=-0.25mm-\tcboxedtitleheight/2, 
        xshift=10mm                          
    },
    boxed title style={
        colback=white,           
        left=0pt, right=0pt, top=0pt, bottom=0pt,
        boxsep=0pt,
    },
    boxsep=5pt,
    left=8pt, right=5pt, top=5pt, bottom=5pt,
}

\algnewcommand{\LeftComment}[1]{\Statex \quad \(\triangleright\) #1}

\renewcommand{\paragraph}[1]{\vspace{0.2cm}\noindent\textbf{#1}}

\newtcbox{\hlprimarytab}{on line, rounded corners, box align=base, colback=green!10,colframe=white,size=fbox,arc=3pt, before upper=\strut, top=-2pt, bottom=-4pt, left=-2pt, right=-2pt, boxrule=0pt}
\newtcbox{\hlsecondarytab}{on line, box align=base, colback=red!10,colframe=white,size=fbox,arc=3pt, before upper=\strut, top=-2pt, bottom=-4pt, left=-2pt, right=-2pt, boxrule=0pt}
\usepackage{mdframed}

\newcommand{\dashifted}{\raisebox{0.5\depth}{\tiny$\downarrow$}}
\newcommand{\uashifted}{\raisebox{0.5\depth}{\tiny$\uparrow$}}

\newcommand{\dar}[1]{{\raisebox{0.6ex}{\tiny\hlsecondarytab{\dashifted{#1}}}}}
\newcommand{\uar}[1]{{\raisebox{0.6ex}{\tiny\hlprimarytab{\uashifted{#1}}}}}

\definecolor{tkcolor}{RGB}{93, 163, 232}
\newtcolorbox{takeaways}[2][]{
    width=\columnwidth,
    toprule=0.0pt,
    leftrule=0.9pt,
    bottomrule=0.9pt,
    rightrule=0.9pt,
    arc=0pt,
    colback = tkcolor!10, 
    colframe = tkcolor, 
    boxsep=0pt,left=7pt,right=7pt,top=4pt,bottom=4pt,
    fontupper=\linespread{0.1}\selectfont,
    title=#2,#1,
    before skip=0.7em,  
    after skip=0.7em
}

\title{Socratic-Zero: Bootstrapping Reasoning via Data-Free Agent Co-evolution}

\author{
  {\bf
    Shaobo Wang $^*$$^{{\color{authorcolor}\boldsymbol{\beta}}}$$\quad$
    Zhengbo Jiao
    $^*$$^{{\color{authorcolor}\boldsymbol{\alpha,\beta,\gamma}}}$$\quad$
    Zifan Zhang
    $^{{\color{authorcolor}\boldsymbol{\alpha,\delta}}}$$\quad$
    Yilang Peng
    $^{{\color{authorcolor}\boldsymbol{\alpha,\epsilon}}}$$\quad$
    \vspace{2pt}
  } \\
  {
  \bf
    Xu Ze
    $^{{\color{authorcolor}\boldsymbol{\alpha}}}$$\quad$
    Boyu Yang
    $^{{\color{authorcolor}\boldsymbol{\alpha}}}$$\quad$    
    Wei Wang
    $^{{\color{authorcolor}\boldsymbol{\alpha}}}$$\quad$
    Hu Wei
    $^{\dagger}$$^{{\color{authorcolor}\boldsymbol{\alpha}}}$$\quad$
    Linfeng Zhang
    $^{\dagger}$$^{{\color{authorcolor}\boldsymbol{\beta}}}$
    \vspace{2pt}
  } \\
    \small
    {
    $^{\color{authorcolor}\boldsymbol{\alpha}}$ Alibaba Group Holding Limited $\quad$
    $^{\color{authorcolor}\boldsymbol{\beta}}$ EPIC Lab, Shanghai Jiao Tong University $\quad$
    \vspace{2pt}
    } \\
    \small
    {
    $^{\color{authorcolor}\boldsymbol{\gamma}}$ Shanghai University of Finance and Economics  
    $^{\color{authorcolor}\boldsymbol{\delta}}$ Wuhan University  $^{\color{authorcolor}\boldsymbol{\epsilon}}$ Zhejiang University
    \vspace{2pt}
    } \\
    \small
    {
    * Equal contribution $\quad$ 
    $^{\dagger}$ Corresponding authors  $\quad$
    }
}

\iclrfinalcopy 

\begin{document}

\maketitle

\vspace{-12pt}
\begin{abstract}
Recent breakthroughs in large language models (LLMs) on reasoning tasks rely heavily on massive, high-quality datasets—typically human-annotated and thus difficult to scale. While data synthesis or distillation offers a promising alternative, existing methods struggle with inconsistent data quality and an inability to dynamically adapt to the evolving capabilities of the model, leading to suboptimal training signals. To address these limitations, we introduce \textit{Socratic-Zero}, a fully autonomous framework that generates high-quality training data from minimal seed examples through the co-evolution of three agents: the \textit{Teacher}, the \textit{Solver}, and the \textit{Generator}. The \textit{Solver} continuously refines its reasoning by learning from preference feedback on both successful and failed trajectories; the \textit{Teacher} adaptively crafts increasingly challenging questions based on the Solver's weaknesses; and the \textit{Generator} distills the Teacher's question-design strategy to enable scalable, high-fidelity curriculum generation. This closed-loop system produces a self-improving curriculum—requiring no pre-existing tasks or labels. Remarkably, starting from only 100 seed questions, our \textit{Socratic-Solver-8B} achieves an average gain of +20.2 percentage points over prior data synthesis methods across seven mathematical reasoning benchmarks (AMC23, AIME24-25, Olympiad, MATH-500, Minerva, and GSM8K), with consistent gains on both Qwen3 and GLM4 series models. Even more surprisingly, synthetic data from \textit{Socratic-Generator-32B} enables student LLMs to achieve superior performance compared to other state-of-the-art (SOTA) commercial LLMs on these benchmarks, including Qwen3-235B-A22B, DeepSeek-V3.1-671B, GPT-5, Gemini-2.5-Pro, Grok-4, and Claude-4.1-Opus.

\end{abstract}

\begin{figure}[h]
\vspace{-20pt}
    \centering
\includegraphics[width=0.99\linewidth]{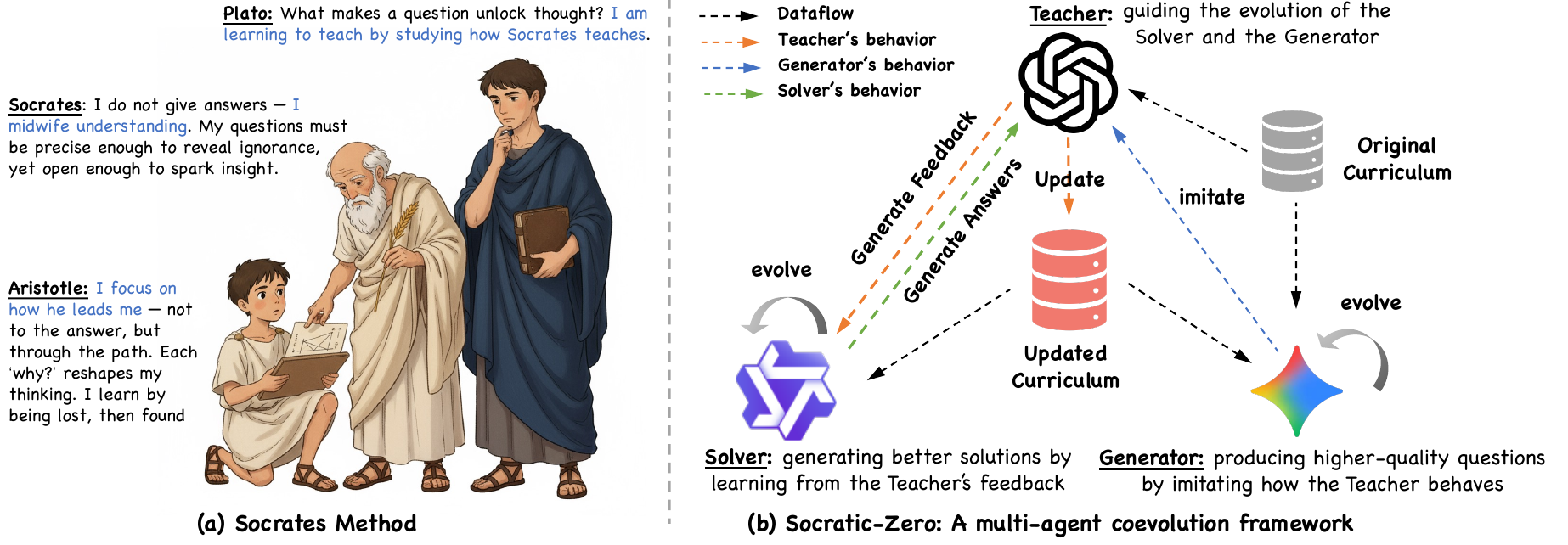}
\vspace{-5pt}
\caption{The Socratic-Zero Framework: From Philosophical Analogy to a Co-evolutionary System.  
\textbf{(a) The Socratic Methodlogy} illustrates the philosophical foundation: the \textbf{Teacher (Socrates)} acts as an intellectual midwife, eliciting understanding through probing questions; the \textbf{Practitioner (Aristotle)} learns not by receiving answers, but by being guided along a path of reasoned inquiry; and the \textbf{Apprentice-Teacher (Plato)} learns to teach by observing and internalizing the master's method.  
\textbf{(b) The Socratic-Zero Framework} operationalizes this philosophy. Here, the \textbf{Teacher}—a powerful LLM—guides the co-evolution of two agents. The \textbf{Solver} improves by generating solutions and refining them through the Teacher's feedback, while the \textbf{Generator}  evolves by strategically distilling the Teacher's behavior to produce an increasingly suitable curriculum for the Solver.}
\vspace{-15pt}
\label{fig:socratic-zero-ab}
\end{figure}

\section{Introduction}
\label{sec:introduction}

The pursuit of advanced mathematical reasoning in large language models has reached a critical juncture. While recent breakthroughs have demonstrated remarkable capabilities on complex mathematical problems~\citep{hendrycks2021measuringmathematicalproblemsolving, cobbe2021trainingverifierssolvemath}, these advances rely on massive datasets of meticulously curated reasoning trajectories — a requirement that is both costly and fundamentally unscalable. Current state-of-the-art models depend on millions of human-annotated problem-solution pairs and hand-designed curricula~\citep{yu2024metamathbootstrapmathematicalquestions}, creating a fundamental bottleneck that limits both accessibility and the potential for models to evolve beyond human-curated knowledge boundaries.

Current methodologies remain entrenched in a static paradigm: datasets are frozen upon collection, curricula are handcrafted in advance, and models are trained on fixed problem distributions. This approach suffers from critical weaknesses: it cannot adapt to evolving model capabilities during training, fails to exploit rich feedback signals for targeting specific weaknesses, and requires extensive human expertise for curriculum design. Recent efforts through synthetic data generation~\citep{lee2024llm2llm, chen2025warriormath} and iterative training~\citep{zhao2025absolute, huang2025rzero} have shown promise but remain constrained by their reliance on external supervision and lack of effective quality control mechanisms for synthesized content.

To overcome these limitations, we introduce \textbf{Socratic-Zero}, a paradigm-shifting framework that eliminates dependency on large-scale external datasets while enabling truly autonomous reasoning improvement. Inspired by the Socratic method of learning through questioning (Figure~\ref{fig:socratic-zero-ab}(a)), our approach implements co-evolution between three agents: a \textit{Solver} that attempts to solve mathematical questions, a \textit{Teacher} that strategically generates challenging problems to expose the Solver's weaknesses, and a \textit{Generator} that learns to distill and scale the Teacher's problem generation strategy. This architecture (Figure~\ref{fig:socratic-zero-ab}(b)) translates the philosophical dialogue of the Socratic method into a concrete, co-evolutionary computational framework. Unlike conventional pipelines that decouple data generation from model training, Socratic-Zero unifies them within a continuous co-evolutionary loop formalized as a optimization problem. Our contributions are threefold:

\begin{figure}[tb!]
    \centering
     \includegraphics[width=0.99\linewidth]{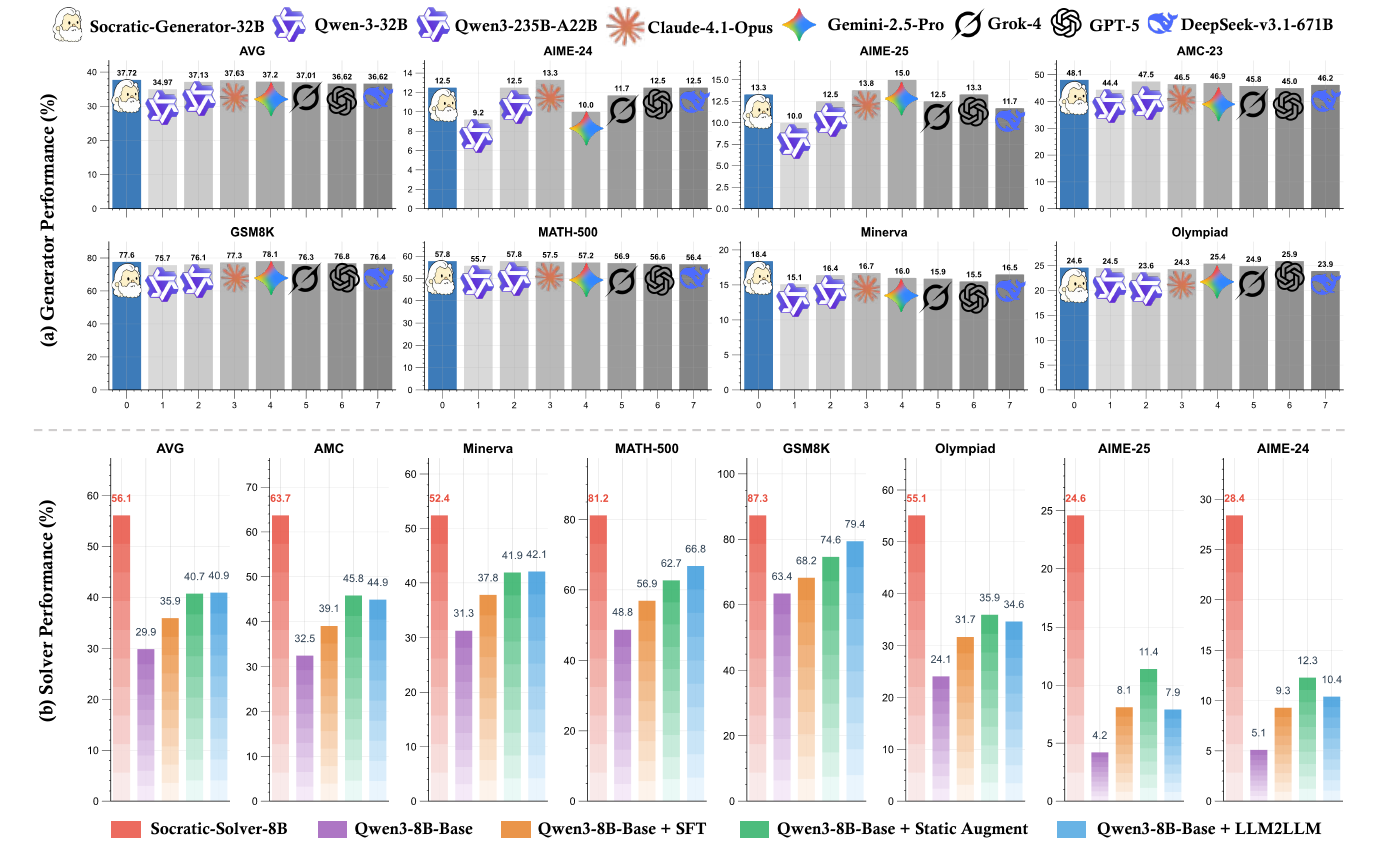}
    \caption{Overall performance comparison demonstrating the giant effectiveness of Socratic-Zero.
    (a) Our Socratic-Generator-32B produces synthetic data that enables student models to achieve performance competitive with much larger state-of-the-art models, showcasing strong generalization capabilities. 
    (b) Our Socratic-Solver-8B achieves an impressive 56.1\% average accuracy, marking a substantial +20.2 point improvement over the baseline. }
    \label{fig:comparison}
\end{figure}

\begin{itemize}[leftmargin=10pt, topsep=0pt, itemsep=1pt, partopsep=1pt, parsep=1pt]
\item \textbf{Multi-Agent Co-Evolutionary Framework:} We establish a theoretical foundation for co-evolutionary learning where the Solver, Teacher, and Generator agents interact dynamically, formalizing reasoning improvement as an adaptive curriculum learning problem (Figure~\ref{fig:socratic-zero-ab}(b)).

\item \textbf{Socratic-Zero System:} We implement a concrete framework where the Solver improves via preference learning, the Teacher evaluates correctness and generates adaptive curriculum, and the Generator learns strategic distillation through value-weighted supervised fine-tuning (WSFT), enabling autonomous reasoning advancement from minimal seed data.

\item \textbf{Superior Empirical Performance:} Our Socratic-Solver-8B achieves +20.2 points average improvement across seven mathematical reasoning benchmarks (Figure~\ref{fig:comparison}(b)), while synthetic data from our Socratic-Generator-32B achieves 37.72\% downstream training effectiveness, outperforming leading commercial models including Qwen3-235B-A22B at 37.13\%, Claude-4.1-Opus at 37.63\%, Gemini-2.5-Pro at 37.20\%, Grok-4 at 37.01\%, GPT-5 at 36.62\%, and DeepSeek-V3.1 at 36.62\% (Figure~\ref{fig:comparison}(a)).

\end{itemize}

\section{Related Work}
\label{sec:related_work}

\paragraph{Data Synthesis.}
To alleviate data scarcity, researchers have leveraged LLMs' generative capabilities to synthesize training samples. Early approaches used prompt engineering to guide question-answer generation~\citep{yu2024metamathbootstrapmathematicalquestions, zhan2025mathsmith}. Subsequently, LLM2LLM~\citep{lee2024llm2llm} and WarriorMath~\citep{chen2025warriormath} introduced deficiency-aware mechanisms, where teacher models identify knowledge gaps and generate targeted data. More recently, Absolute Zero~\citep{zhao2025absolute} and R-Zero~\citep{huang2025rzero} explored fully autonomous self-play paradigms for continuous task generation and learning. While these advances achieve data autonomy, they lack effective quality control mechanisms, resulting in repeated use of low-value samples that severely impact effectiveness.

\paragraph{Data Distillation.}
Knowledge distillation transfers capabilities from powerful teacher models to lighter student models. Early work like Orca~\citep{mukherjee2023orca} used imitation learning to replicate teacher reasoning chains. Policy distillation~\citep{wang2025rlkd} extends this by transferring dynamic decision-making strategies. GKD~\citep{agarwal2024onpolicy} enables students to learn from their own sequences using teacher feedback for policy correction. However, students passively accept teacher feedback without evaluating reliability, degrading learning quality when guidance is suboptimal. These methods also rely on static datasets, unable to dynamically adjust content based on students' evolving capabilities. While recent advances~\citep{Wang_2025_CVPR,zhao2023data,zhang2024dilm,chen2024distillm,liu2025shiftingaiefficiency} promote data-centric optimization, they lack effective quality control and adaptive curriculum generation.

\paragraph{Preference Learning.}
Translating feedback signals into model optimization is central to self-evolution systems. Early approaches like RLHF~\citep{stiennon2020learning} trained reward models on human preferences then fine-tuned policies, but this process is complex and unstable. Recent methods like DPO~\citep{rafailov2023direct} and RWSFT~\citep{mukherjee2025learning} directly optimize preferences, improving efficiency and stability. Combined with self-correction mechanisms like Self-Refine~\citep{madaan2023selfrefine}, models possess preliminary closed-loop capabilities. Further advances including Self-Evolved Reward Learning~\citep{huang2025selfevolved}, Self-Play Fine-Tuning~\citep{chen2024selfplay}, and Self-Play Critic~\citep{chen2025spc} explore autonomous feedback strategies. However, these methods lack unified, co-evolving frameworks for feedback generation and validation.


\section{Methodology}
\label{sec:methodology}

\subsection{The Socratic-Zero Framework}
\label{sec:framework}

We introduce \textbf{Socratic-Zero}, a fully autonomous, co-evolutionary framework designed to bootstrap mathematical reasoning from a minimal set of seed problems, entirely without relying on external human-annotated data. As illustrated in Figure~\ref{fig:pipeline}, the system operates as a self-improving loop among three agents: a \textbf{Solver} that learns to reason, a fixed \textbf{Teacher} that acts as an oracle for evaluation and problem refinement, and a \textbf{Generator} that learns to synthesize a curriculum.

At each iteration $t$, the framework operates on a curriculum of problems and their reference solutions, denoted as $\mathcal{D}_t = \{(q, y_{\text{ref}})\}$. The core of Socratic-Zero is the co-evolution of the Solver and Generator under the guidance of the Teacher. The Solver is trained to solve problems from $\mathcal{D}_t$. Its failures are then used by the Teacher to create new, targeted problems. The Generator, in turn, distills the Teacher's strategy to produce a scalable curriculum. This process creates a dynamic curriculum that continuously adapts to the Solver's evolving capabilities, ensuring the training signals remain maximally informative. The agents are formally defined as follows:

\begin{figure}[tb!]
    \centering
    \includegraphics[width=0.99\linewidth]{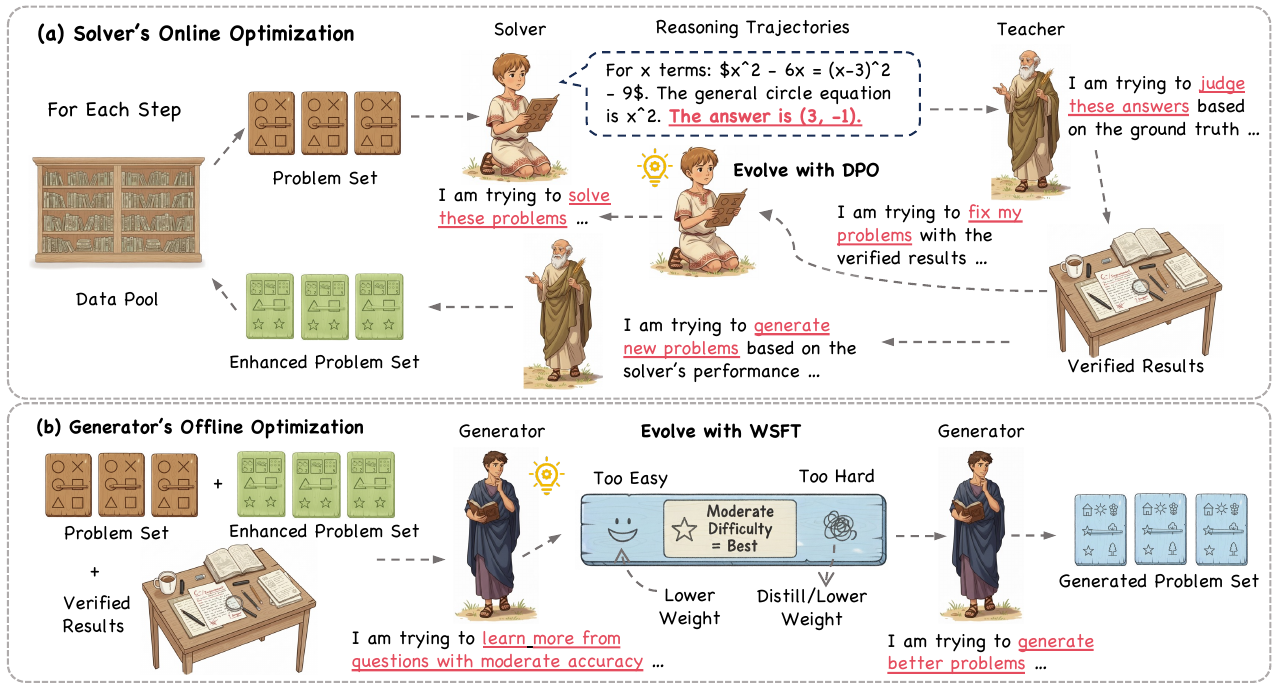}
\caption{Overview of the Socratic-Zero Framework. (a) Solver Evolving: The Solver attempts to solve problems and learns from preference pairs of correct and incorrect solutions via DPO, while the frozen Teacher strategically generates challenging problems based on Solver failures using fixed generation and evaluation functions. (b) Generator Evolving: The Generator distills the Teacher's problem generation strategy using value-weighted supervised learning. Together, these create a self-improving loop where the curriculum dynamically evolves to maintain optimal challenge levels for the Solver's current capabilities.}
    \label{fig:pipeline}
\end{figure}

\begin{tcolorbox}[colframe=black!75!gray,title=Agents in the Socratic-Zero Framework,boxsep=3pt,left=3pt,right=3pt,top=3pt,bottom=3pt]
\begin{enumerate}[leftmargin=10pt, topsep=0pt, itemsep=1pt, partopsep=1pt, parsep=1pt]
    \item \textbf{Solver ($\mathcal{S}$):} An agent with a policy $\pi_{\theta_{\mathcal{S}}}$, parameterized by $\theta_{\mathcal{S}}$, which maps a problem $q$ to a solution trajectory $y$. Its objective is to generate correct reasoning paths. At each iteration $t$, it improves by learning from preference feedback on its own attempts.

    \item \textbf{Teacher ($\mathcal{T}$):} A fixed, high-capacity LLM that provides two deterministic oracle functions:
    (i) a verification function $V(q, y) \to \{0, 1\}$, which judges the correctness of a solution $y$ for a problem $q$; and
    (ii) a problem refinement function $G(q, y_{\text{fail}}) \to (q', y'_{\text{ref}})$, which creates a new problem-solution pair by revising an original problem $q$ based on a failed solution $y_{\text{fail}}$.

    \item \textbf{Generator ($\mathcal{G}$):} An agent with a policy $\pi_{\theta_{\mathcal{G}}}$, parameterized by $\theta_{\mathcal{G}}$, that learns to mimic the Teacher's refinement strategy. It maps a problem and a failed solution $(q, y_{\text{fail}})$ to a new problem $q'$. It evolves to generate problems that are optimally challenging for the current Solver.
\end{enumerate}
\end{tcolorbox}

The curriculum expands based on the Solver's mistakes. At iteration $t$, the set of Solver failures on the current curriculum $\mathcal{D}_t$ is collected:
\begin{equation}
\mathcal{F}_t = \left\{ (q, y_{\mathcal{S}}) \mid (q, y_{\text{ref}}) \in \mathcal{D}_t, y_{\mathcal{S}} \sim \pi_{\theta_{\mathcal{S}}^{(t)}}(\cdot \mid q), V(q, y_{\mathcal{S}}) = 0 \right\}.
\end{equation}
The Teacher refines each failure into a new, instructive problem-solution pair. The set of these new pairs, $\mathcal{D}_{\text{new}}$, is used to augment the curriculum for the next iteration:
\begin{equation}
\label{eq:curriculum_update}
\mathcal{D}_{t+1} = \mathcal{D}_t \cup \underbrace{\left\{ G(q, y_{\mathcal{S}}) \mid (q, y_{\mathcal{S}}) \in \mathcal{F}_t \right\}}_{\mathcal{D}_{\text{new}}}.
\end{equation}
The full co-evolutionary training procedure is detailed in Algorithm~\ref{alg:socratic}.

\subsection{Solver Training via Online Preference Optimization}
\label{sec:solver_advancement}

The Solver's policy $\pi_{\theta_{\mathcal{S}}}$ is improved through online preference learning, leveraging the Teacher's verification function $V$ to create a feedback loop. For each problem $q \in \mathcal{D}_t$, the Solver generates $k$ solution attempts, $\{y_{\mathcal{S}}^{(i)}\}_{i=1}^k$. These attempts are partitioned into a set of ``winning'' (correct) solutions $\mathcal{Y}_w(q)$ and ``losing'' (incorrect) solutions $\mathcal{Y}_l(q)$:
\begin{align}
\mathcal{Y}_l(q) &= \{ y_{\mathcal{S}}^{(i)} \mid V(q, y_{\mathcal{S}}^{(i)}) = 0 \}, \\
\mathcal{Y}_w(q) &= \{ y_{\mathcal{S}}^{(i)} \mid V(q, y_{\mathcal{S}}^{(i)}) = 1 \} \cup \{y_{\text{ref}} \mid \text{if } \forall i, V(q, y_{\mathcal{S}}^{(i)}) = 0\},
\end{align}
where $y_{\text{ref}}$ is the reference solution from the curriculum. If the Solver fails to generate any correct solution, the ground-truth solution serves as the sole winning example. This ensures a valid preference pair $(y_w, y_l)$, where $y_w \in \mathcal{Y}_w(q)$ and $y_l \in \mathcal{Y}_l(q)$, can always be constructed.

The Solver's parameters $\theta_{\mathcal{S}}$ are then updated using the Direct Preference Optimization (DPO) loss~\citep{rafailov2023direct}. This objective maximizes the likelihood of preferred solutions over rejected ones:
\begin{equation}
\label{eq:dpo_loss}
\mathcal{L}_{\text{DPO}}(\theta_{\mathcal{S}}; \theta_{\text{ref}}) = - \mathbb{E}_{q \sim \mathcal{D}_t, y_w \sim \mathcal{Y}_w(q), y_l \sim \mathcal{Y}_l(q)} \left[ \log \sigma \left( \beta \log \frac{\pi_{\theta_{\mathcal{S}}}(y_w \mid q)}{\pi_{\theta_{\text{ref}}}(y_w \mid q)} - \beta \log \frac{\pi_{\theta_{\mathcal{S}}}(y_l \mid q)}{\pi_{\theta_{\text{ref}}}(y_l \mid q)} \right) \right],
\end{equation}
where $\pi_{\theta_{\text{ref}}}$ is a frozen reference policy (e.g., $\pi_{\theta_{\mathcal{S}}}$ from the start of the iteration), $\beta$ is a temperature hyperparameter, and $\sigma$ is the sigmoid function.

\subsection{Generator Training via Offline Value-Weighted Distillation}
\label{sec:generator_advancement}

To ensure scalable curriculum generation without perpetual reliance on the expensive Teacher, the Generator $\pi_{\theta_{\mathcal{G}}}$ is trained to distill the Teacher's problem refinement strategy. An effective curriculum should feature problems of \textit{desirable difficulty}—challenging enough to be informative but not so difficult as to be unsolvable.

We formalize this concept with a utility function $U(q' | \pi_{\theta_{\mathcal{S}}})$ that scores a new problem $q'$ based on the current Solver's performance. Let $s_q = \frac{1}{k} \sum_{i=1}^k V(q, y_{\mathcal{S}}^{(i)})$ be the success rate of the Solver $\pi_{\theta_{\mathcal{S}}}$ over $k$ attempts. The utility is defined by an unnormalized Gaussian centered at a target success rate $\mu$:
\begin{equation}
\label{eq:utility_function}
U(q' | \pi_{\theta_{\mathcal{S}}}) = \exp\left( -\frac{(s_{q'} - \mu)^2}{2\sigma^2} \right).
\end{equation}
We set $\mu=0.5$ to incentivize problems at the frontier of the Solver's capabilities, with $\sigma$ controlling the tolerance for deviation.

The Generator is trained via weighted supervised fine-tuning (WSFT) to mimic the Teacher's generation of high-utility problems. The training data $\mathcal{D}_{\mathcal{G}}$ is constructed from the set of Solver failures $\mathcal{F}_t$ and the corresponding Teacher-refined problems from $\mathcal{D}_{\text{new}}$:
$$ \mathcal{D}_{\mathcal{G}} = \{ (q, y_{\text{fail}}, q') \mid (q, y_{\text{fail}}) \in \mathcal{F}_t, (q', y'_{\text{ref}}) = G(q, y_{\text{fail}}) \}. $$
The Generator's objective is to maximize the utility-weighted log-likelihood of producing the Teacher's refined problems:
\begin{equation}
\label{eq:generator_loss}
\mathcal{L}_{\text{WSFT}}(\theta_{\mathcal{G}}) = - \mathbb{E}_{(q, y_{\text{fail}}, q') \sim \mathcal{D}_{\mathcal{G}}} \left[ U(q' | \pi_{\theta_{\mathcal{S}}}) \cdot \log \pi_{\theta_{\mathcal{G}}}(q' \mid q, y_{\text{fail}}) \right].
\end{equation}
This objective steers the Generator towards producing problems that are optimally challenging for the current Solver, effectively internalizing the Teacher's expert curriculum design principles.

\section{Experiments}
\label{sec:experiments}

\subsection{Experimental Setup}
\label{subsec:setup}

\textbf{Models.}
We employed Qwen3-235B-A22B-Instruct-2507~\citep{yang2025qwen3technicalreport} as the Teacher model to provide high-quality evaluation and curriculum generation. We used Qwen3-32B~\citep{yang2025qwen3technicalreport} as the Generator to learn and distill the Teacher's problem generation strategies. We conducted Solver experiments on multiple model architectures including Qwen3-8B-base, Qwen3-14B-base~\citep{yang2025qwen3technicalreport}, and GLM4-9B-base~\citep{glm2024chatglmfamilylargelanguage} to demonstrate cross-model generalization. We compared strong baselines including Gemini-2.5-Pro-06-17~\citep{comanici2025gemini25pushingfrontier}, GPT5-0807-global, and DeepSeek-v3.1-671B~\citep{deepseekai2025deepseekv3technicalreport} against our approach. For downstream evaluation of generated data quality, we fine-tuned DeepSeek-R1-Distill-Llama-8B~\citep{deepseekai2025deepseekr1incentivizingreasoningcapability} as the student model.

\textbf{Benchmarks.}
We used seven mathematical reasoning benchmarks for evaluation, including AMC~\citep{cao2025stepguidedreasoningimproving}, Minerva~\citep{nagrani2025minervaevaluatingcomplexvideo}, MATH-500~\citep{hendrycks2021measuringmathematicalproblemsolving}, GSM8K~\citep{cobbe2021trainingverifierssolvemath}, Olympiad-Bench~\citep{he2024olympiadbenchchallengingbenchmarkpromoting}, AIME-2024, and AIME-2025. Additionally, we employed three general reasoning benchmarks to assess the transfer of mathematical reasoning improvements to broader cognitive abilities, namely BBEH~\citep{kazemi2025bigbenchextrahard}, MMLU-Pro~\citep{wang2024mmluprorobustchallengingmultitask}, and SuperGPQA~\citep{pteam2025supergpqascalingllmevaluation}.

\textbf{Curriculum Settings.}
The initial curriculum $\mathcal{C}_0$ contained 100 questions sampled from the MATH training set~\citep{hendrycks2021measuringmathematicalproblemsolving} following specific diversity and difficulty criteria (detailed in Appendix~\ref{subsec:seed_selection}). All Solver models first underwent LoRA-based~\citep{hu2021lora} SFT on a 1,500-problem dataset of Level 5 difficulty. Key hyperparameters: $k=8$ solution trajectories per problem,  reward parameters $\mu=0.5$ and $\sigma=0.2$, and training batches combined 100\% new problems with 25\% historical curriculum for replay.

\textbf{Solver Evaluation.}
For each test question, we generated 32 solutions using zero-shot prompting with temperature 0.7. We determined correctness through a dual-verification mechanism combining rule-based answer extraction and semantic validation. We reported Mean@32 accuracy across all evaluations. Detailed evaluation protocols, including sampling strategies, answer extraction methods, and LLM judge configurations, are provided in Appendix~\ref{subsec:evaluation_protocol}.

\textbf{Baselines.}
\textbf{Baselines.}
We employed two strong baselines for comparison. \textit{Static Augmentation (SA)} follows traditional approaches via MetaMath~\citep{yu2024metamathbootstrapmathematicalquestions} and WizardMath~\citep{luo2023wizardmathempoweringmathematicalreasoning}, augmenting training data with fixed synthetic questions generated offline without adaptive curriculum evolution. \textit{LLM2LLM}~\citep{lee2024llm2llm} implements iterative self-training where models generate questions based on current failures and retrain on augmented datasets. Both baselines use identical SFT initialization for fair comparison.

\textbf{Generator Evaluation.}
We prompted each generator with 1,000 SAND-Math~\citep{zhang2025sandmath} seeds to produce 3 variants each, resulting in 3,000 total generated questions. We measured validity rate by having Qwen3-235B-A22B-Instruct-2507 attempt to solve each generated question under a 4,096-token, 600-s timeout constraint. We evaluated downstream utility by fine-tuning DeepSeek-R1-Distill-Llama-8B on the QAs and measuring performance on mathematical reasoning benchmarks.

\textbf{Infrastructure.}
We conducted training experiments on 8×NVIDIA H20 GPUs. We performed Teacher model inference using 16×AMD MI308X GPUs. Detail provided in Appendix~\ref{subsec:teacher_infrastructure}.

\subsection{Solver Results}
\label{subsec:solver_results}

\paragraph{Baseline Comparison.}
Table~\ref{tab:solver_main_results_absolute} shows Socratic achieves 56.1\% average accuracy, outperforming Static Augmentation by +15.4 points and LLM2LLM by +15.2 points. Notable gains appear on competition problems: AIME-24 (+19.1) and AIME-25 (+16.5), demonstrating the advantages of DPO-based preference learning and adaptive curriculum generation.

\paragraph{Cross-Architecture Generalization.}
Table~\ref{tab:cross_model_generalization} validates that Socratic principles transcend specific model families. On GLM4-9B-base, Socratic Stage 3 achieves 52.3\% average accuracy (+17.1 points over base model), with strong improvements on AIME benchmarks: AIME-25 (+20.4) and AIME-24 (+23.9). Similarly, on Qwen3-14B-base, Stage 3 reaches 60.3\% (+17.3 points), demonstrating consistent effectiveness across different architectures and addressing fundamental reasoning capabilities.

\paragraph{Transfer to General Reasoning.}
Table~\ref{tab:general_reasoning_results_reformatted} shows mathematical reasoning improvements transfer to broader cognitive abilities, with +6.02 points average improvement across BBEH, MMLU-Pro, and SuperGPQA benchmarks.

\begin{table}[tb!]
\centering
\caption{Solver Evaluation Results with different training methods. Results are reported on seven benchmarks (AMC, Minerva, MATH-500, GSM8K, Olympiad, AIME-25, AIME-24) and their average. Arrow values represent absolute point changes relative to Static Augmentation, where $\uparrow$ indicates improvement and $\downarrow$ indicates decline.}
\label{tab:solver_main_results_absolute}
\resizebox{0.99\textwidth}{!}{%
\begin{tabular}{l|ccccccc|c}
\toprule
\multirow{2}{*}{\textbf{Training Method}} & \multicolumn{7}{c|}{\textbf{Benchmark Datasets}} & \multirow{2}{*}{\textbf{Avg.}} \\
 & AMC & Minerva & MATH-500 & GSM8K & Olympiad & AIME-25 & AIME-24 & \\
\midrule

\multicolumn{9}{l}{\textit{Qwen3-8B-base}} \\
\quad + Zero-shot & 32.5 & 31.3 & 48.8 & 63.4 & 24.1 & 4.2 & 5.1 & 29.9 \\
\quad + SFT  & 39.1 & 37.8 & 56.9 & 68.2 & 31.7 & 8.1 & 9.3 & 35.9 \\
\midrule

\quad + Static Augmentation & 45.8 & 41.9 & 62.7 & 74.6 & 35.9 & 11.4 & 12.3 & 40.7 \\
\midrule

\multicolumn{9}{l}{\textit{Qwen3-8B-base with LLM2LLM}} \\
\quad + Stage 1 & 41.6\dar{4.2} & 41.2\dar{0.7} & 53.1\dar{9.6} & 78.3\uar{3.7} & 32.4\dar{3.5} & 6.7\dar{4.7} & 8.9\dar{3.4} & 37.5\dar{3.2} \\
\quad + Stage 2 & 43.2\dar{2.6} & 40.6\dar{1.3} & 54.9\dar{7.8} & 79.1\uar{4.5} & 33.8\dar{2.1} & 7.2\dar{4.2} & 9.1\dar{3.2} & 38.3\dar{2.4} \\
\quad + Stage 3 & 44.9\dar{0.9} & 42.1\uar{0.2} & 66.8\uar{4.1} & 79.4\uar{4.8} & 34.6\dar{1.3} & 7.9\dar{3.5} & 10.4\dar{1.9} & 40.9\uar{0.2} \\
\midrule
\multicolumn{9}{l}{\textit{Qwen3-8B-base with Socratic-Zero (Ours)}} \\
\quad + Stage 1 & 43.8\dar{2.0} & 39.4\dar{2.5} & 60.2\dar{2.5} & 69.7\dar{4.9} & 35.3\dar{0.6} & 10.6\dar{0.8} & 11.8\dar{0.5} & 38.7\dar{2.0} \\
\quad + Stage 2 & 49.3\uar{3.5} & 40.7\dar{1.2} & 63.4\uar{0.7} & 71.8\dar{2.8} & 38.2\uar{2.3} & 12.9\uar{1.5} & 15.6\uar{3.3} & 41.7\uar{1.0} \\
\quad + Stage 3
  & \textbf{63.7\uar{17.9}} & \textbf{52.4\uar{10.5}} & \textbf{81.2\uar{18.5}} & \textbf{87.3\uar{12.7}}
  & \textbf{55.1\uar{19.2}} & \textbf{24.6\uar{13.2}} & \textbf{28.4\uar{16.1}} & \textbf{56.1\uar{15.4}} \\
\bottomrule

\end{tabular}%
}
\end{table}

\begin{table}[tb!]
\centering
\caption{Cross-Model Generalization Results with different training stages. Each block corresponds to a model (GLM4-9B, Qwen3-14B). Results are reported on seven benchmarks (AMC, Minerva, MATH-500, GSM8K, Olympiad, AIME-25, AIME-24) and their average. Values with arrows represent absolute point changes relative to SFT for each model.}
\label{tab:cross_model_generalization}
\resizebox{0.99\textwidth}{!}{%
\begin{tabular}{l|ccccccc|c}
\toprule
\multirow{2}{*}{\textbf{Training Method}} & \multicolumn{7}{c|}{\textbf{Benchmark Datasets}} & \multirow{2}{*}{\textbf{Avg}} \\
 & AMC & Minerva & MATH-500 & GSM8K & Olympiad & AIME-25 & AIME-24 & \\
\midrule

\multicolumn{9}{l}{\textit{GLM4-9B-base}} \\
\quad +Zero-shot & 34.5 & 37.3 & 52.3 & 72.5 & 34.8 & 7.5 & 7.2 & 35.2 \\
\quad + SFT & 38.4 & 44.8 & 63.8 & 77.2 & 41.3 & 15.1 & 19.3 & 42.8 \\
\quad + Socratic Stage 1 & 39.4\uar{1.0} & 47.3\uar{2.5} & 67.9\uar{4.1} & 79.8\uar{2.6} & 43.4\uar{2.1} & 15.6\uar{0.5} & 24.0\uar{4.7} & 45.3\uar{2.5} \\
\quad + Socratic Stage 2 & 42.3\uar{3.9} & 49.4\uar{4.6} & 68.1\uar{4.3} & 82.5\uar{5.3} & 45.5\uar{4.2} & 19.1\uar{4.0} & 25.3\uar{6.0} & 47.5\uar{4.7} \\
\quad + Socratic Stage 3 & \textbf{47.5\uar{8.7}} & \textbf{52.8\uar{8.0}} & \textbf{73.8\uar{10.0}} & \textbf{83.9\uar{6.7}} & \textbf{49.4\uar{8.1}} & \textbf{27.9\uar{12.8}} & \textbf{31.1\uar{11.8}} & \textbf{52.3\uar{9.5}} \\
\midrule

\multicolumn{9}{l}{\textit{Qwen3-14B-base}} \\
\quad +Zero-shot & 48.8 & 40.5 & 62.0 & 91.5 & 38.4 & 9.6 & 10.0 & 43.0 \\
\quad + SFT & 61.3 & 51.8 & 71.5 & 92.2 & 47.3 & 18.1 & 20.3 & 51.8 \\
\quad + Socratic Stage 1 & 62.9\uar{1.6} & 55.1\uar{3.3} & 74.6\uar{3.1} & 91.8\dar{0.4} & 52.5\uar{5.2} & 19.8\uar{1.7} & 21.7\uar{1.4} & 54.1\uar{2.3} \\
\quad + Socratic Stage 2 & 65.4\uar{4.1} & 57.4\uar{5.6} & 76.7\uar{5.2} & 92.3\uar{0.1} & 54.2\uar{6.9} & 24.8\uar{6.7} & 23.3\uar{3.0} & 56.3\uar{4.5} \\
\quad + Socratic Stage 3 & \textbf{70.0\uar{8.7}} & \textbf{60.7\uar{8.9}} & \textbf{80.2\uar{8.7}} & \textbf{93.7\uar{1.5}} & \textbf{58.3\uar{11.0}} & \textbf{28.9\uar{10.8}} & \textbf{30.1\uar{9.8}} & \textbf{60.3\uar{8.5}} \\
\bottomrule

\end{tabular}%
}
\end{table}

\begin{table}[tb!]
\centering
\caption{Performance on general reasoning benchmarks with different training stages. Results are reported on three benchmarks (BBEH, MMLU-Pro, SuperGPQA) and their average. Values with arrows represent absolute point changes relative to zero-shot Qwen3-8B-base performance.}
\vspace{-5pt}
\label{tab:general_reasoning_results_reformatted}
{\small
\begin{tabular}{l|ccc|c}
\toprule 
\multirow{2}{*}{\textbf{Training Method}} & \multicolumn{3}{c|}{\textbf{General Reasoning Benchmarks}} & \multirow{2}{*}{\textbf{Avg.}} \\
 & BBEH & MMLU-Pro & SuperGPQA & \\ 
\midrule
\multicolumn{5}{l}{\textit{Qwen3-8B-Base}} \\
\quad + Zero-shot & 7.68 & 50.00 & 24.73 & 27.47 \\
\midrule
\multicolumn{5}{l}{\textit{Base Model with Socratic (Ours)}} \\
\quad + Stage 1 & 8.48\uar{0.80} & 55.71\uar{5.71} & 27.32\uar{2.59} & 30.50\uar{3.03} \\
\quad + Stage 2 & 9.11\uar{1.43} & 59.29\uar{9.29} & 29.73\uar{5.00} & 32.71\uar{5.24} \\
\quad \textbf{+ Stage 3} 
  & \textbf{9.54\uar{1.86}} & \textbf{60.89\uar{10.89}} & \textbf{30.05\uar{5.32}} & \textbf{33.49\uar{6.02}} \\
\bottomrule
\end{tabular}%
}
\vspace{-10pt}
\end{table}

\subsection{Generator Results}
\label{subsec:generator_results}

We assessed both the intrinsic quality of generated problems and their downstream training effectiveness, with Socratic-Generator-32B being compared against its base model and SOTA commercial large language models to determine whether strategic specialization can match the performance of much advanced larger models.

\subsubsection{Evaluation Protocol}

We adopted a standardized, three-stage evaluation pipeline to holistically assess both the \textit{intrinsic quality} of generated problems and their \textit{extrinsic utility} in downstream model training. The full procedure is formalized below.

\textbf{Step 1: Problem Generation.} We prompted each generator with 1,000 seed problems from SAND-Math~\citep{zhang2025sandmath} and tasked with producing five augmented variants per seed, resulting in 3,000 total generated problems per model.

\textbf{Step 2: Quality Assessment.} We measured problem validity by prompting Qwen3-235B~\citep{yang2025qwen3technicalreport} — selected for its state-of-the-art mathematical reasoning capability and its role as the teacher model in the distillation framework — to solve each generated problem under strict constraints: a 4,096-token limit and a 600-second timeout. The \textit{Validity Rate} was defined as the percentage of problems successfully solved within these bounds.

\textbf{Step 3: Student Evaluation.} We used all valid question-answer (QA) pairs to fine-tune the student model, DeepSeek-R1-Distill-Llama-8B~\citep{deepseekai2025deepseekr1incentivizingreasoningcapability}. We evaluated \textit{Downstream Utility} as the mean accuracy — average accuracy over 16 independent decoding runs per problem — across seven diverse mathematical reasoning benchmarks.

\subsubsection{Problem Quality Assessment}

\begin{wraptable}{r}{0.46\textwidth}
\vspace{-15pt}
\caption{Generator validity rates.}
\label{tab:generator_quality}
\centering
\vspace{-10pt}
\small
\begin{tabular}{lc}
\toprule
\textbf{Generator Model} & \textbf{Validity Rate (\%)} \\
\midrule
Qwen3-32B & 89.1 \\
Qwen3-235B-A22B & 95.1\uar{6.0} \\
Gemini-2.5-Pro & 94.2\uar{5.1} \\
GPT5-global & 95.8\uar{6.7} \\
DeepSeek-v3.1-671B & 96.5\uar{7.4} \\
Grok4& 95.7\uar{6.7} \\
Claude-4.1-opus& 96.9\uar{7.} \\
\midrule
\textbf{Socratic-Generator-32B} & \textbf{95.6\uar{6.5}} \\
\bottomrule
\end{tabular}
\end{wraptable}
To evaluate the quality of the generated problems, we measure their Validity Rate — the percentage of problems solvable by a powerful model (Qwen3-235B-A22B-Instruct-2507). As shown in Table~\ref{tab:generator_quality}, our specialized Socratic-Generator-32B generator achieves a remarkable 95.6\% validity rate. This not only represents a substantial improvement over its base model Qwen3-32B but also rivals the performance of significantly larger models, including proprietary models like GPT5-0807-global, Gemini-2.5-Pro-06-17. This demonstrates our co-evolutionary strategy effectively.

\subsubsection{Downstream Effectiveness}

Table~\ref{tab:generator_utility} reports the downstream utility of each generator, measured by the performance of the fine-  student model. The output from our Socratic-Generator-32B leads to a final student accuracy of 37.72\%. Notably, this performance not only rivals that achieved using data from significantly larger models but also marginally surpasses (+0.59 points) the result from its own Teacher (Qwen3-235B), despite being over 20x smaller.

\begin{table}[tb!]
\centering
\caption{Downstream Training Effectiveness with different generator models. Results are reported on seven benchmarks (AIME-24, AIME-25, AMC-23, GSM8K, MATH-500, Minerva, Olympiad) and their average. Values with arrows represent absolute point changes relative to Qwen3-32B baseline.}
\label{tab:generator_utility}
\resizebox{0.99\textwidth}{!}{%
\begin{tabular}{l|ccccccc|c}
\toprule 
\multirow{2}{*}{\textbf{}} & \multicolumn{7}{c|}{\textbf{Benchmark Datasets}} & \multirow{2}{*}{\textbf{Avg.}} \\
 & AIME-24 & AIME-25 & AMC-23 & GSM8K & MATH-500 & Minerva & Olympiad & \\
\midrule

\multicolumn{9}{l}{\textit{DeepSeek-R1-Distill-Llama-8B}} \\
\quad + Zero-shot & 5.8 & 8.3 & 42.5 & 72.2 & 52.4 & 15.3 & 23.0 & 32.75 \\
\midrule

\multicolumn{9}{l}{\textit{Open-Sourced Generators}} \\
\quad  Qwen3-32B & 9.2 & 10.0 & 44.4 & 75.7 & 55.7 & 15.1 & 24.5 & 34.97 \\
\quad  Qwen3-235B-A22B-Instruct-2507 & 12.5\uar{3.3} & 12.5\uar{2.5} & 47.5\uar{3.1} & 76.1\uar{0.4} & 57.8\uar{2.1} & 16.4\uar{1.3} & 23.6\dar{0.9} & 37.13\uar{2.16} \\
\quad  DeepSeek-v3.1-671B & 12.5\uar{3.3} & 11.7\uar{1.7} & 46.2\uar{1.8} & 76.4\uar{0.7} & 56.4\uar{0.7} & 16.5\uar{1.4} & 23.9\dar{0.6} & 36.62\uar{1.65} \\
\midrule

\multicolumn{9}{l}{\textit{Advanced commercial Generators}} \\

\quad  Gemini-2.5-Pro-06-17 & 10.0\uar{0.8} & 15.0\uar{5.0} & 46.9\uar{2.5} & 78.1\uar{2.4} & 57.2\uar{1.5} & 16.0\uar{0.9} & 25.4\uar{0.9} & 37.20\uar{2.23} \\
\quad  GPT5-0807-global & 12.5\uar{3.3} & 13.3\uar{3.3} & 45.0\uar{0.6} & 76.8\uar{1.1} & 56.6\uar{0.9} & 15.5\uar{0.4} & 25.9\uar{1.4} & 36.62\uar{1.65} \\
\quad  Grok-4 & 11.7\uar{2.5} & 12.5\uar{2.5} & 45.8\uar{1.4} & 76.3\uar{0.6} & 56.9\uar{1.2} & 15.9\uar{0.8} & 24.9\uar{0.4} & 37.01\uar{2.04} \\
\quad  Claude-4.1-Opus & 13.3\uar{4.1} & 13.8\uar{3.8} & 46.5\uar{2.1} & 77.3\uar{1.6} & 57.5\uar{1.8} & 16.7\uar{1.6} & 24.3\dar{0.2} & 37.63\uar{2.66} \\

\midrule

\quad  \textbf{Socratic-Generator-32B} 
  & \textbf{12.5\uar{3.3}} & \textbf{13.3\uar{3.3}} & \textbf{48.1\uar{3.7}} & \textbf{77.6\uar{1.9}} 
  & \textbf{57.8\uar{2.1}} & \textbf{18.4\uar{3.3}} & \textbf{24.6\uar{0.1}} & \textbf{37.72\uar{2.75}} \\
\bottomrule

\end{tabular}%
}
\vspace{-5pt}
\end{table}

\subsection{Ablation Studies}
\label{subsec:ablation}

We conducted two key ablation studies to validate our framework's design choices, with results summarized in Table~\ref{tab:ablation_studies}. The first study examines the necessity of initial supervised fine-tuning (SFT), while the second investigates different reward function formulations during reinforcement learning.

\begin{table*}[tb!]
\centering
\caption{Ablation studies on the necessity of initial SFT and different strategies of reward functions. (a) Values with arrows represent absolute point changes relative to the previous stage within the same method. (b) Values with arrows represent absolute point changes relative to the Gaussian baseline. $\rho$ represents solver success rate, $\mu$ represents target success rate, $\sigma$ represents standard deviation in Gaussian reward function $\mathcal{N}(\mu, \sigma)$, $\Psi_\rho(a, b)$ represents linear function $\Psi_\rho(a, b) = a\rho + b$.}
\label{tab:ablation_studies}
\begin{subtable}[t]{0.4\textwidth}
    \centering
    \caption{Ablation Study on Initial SFT (AIME-24)}
    \label{tab:sft_ablation_results}
    \vspace{-5pt}
    \setlength{\tabcolsep}{3pt}
    \renewcommand{\arraystretch}{0.9}
    \footnotesize
    \begin{tabular}{lcc}
    \toprule 
    \textbf{Method} & \textbf{Score (\%)} & \textbf{$\Delta$ (\%)} \\
    \midrule
     \multirow{1}{*}{\textit}  Qwen3-8B-Base & 9.64 & \textbf{-} \\
    \midrule
    \multicolumn{3}{l}{\textit{Socratic-Zero (w/o SFT)}} \\
    \quad + Stage 1 & 11.67 & \uar{2.03} \\
    \quad + Stage 2 & 11.15 & \uar{1.51} \\
    \quad + Stage 3 & 11.98 & \uar{2.34} \\
    \midrule
    \multicolumn{3}{l}{\textit{Socratic-Zero (w/ SFT)}} \\
    \quad + Stage 1 & 13.44 & \uar{3.80} \\
    \quad + Stage 2 & 14.48 & \uar{4.84} \\
    \quad \textbf{+ Stage 3} & \textbf{28.02} & \textbf{\uar{18.38}} \\
    \bottomrule
    \end{tabular}
\end{subtable}%
\hfill 
\begin{subtable}[t]{0.59\textwidth}
    \centering
    \caption{Ablation of Reward Functions (Benchmark Avg.)}
    \vspace{-5pt}
    \label{tab:reward_ablation_results}
    \setlength{\tabcolsep}{2pt}
    \renewcommand{\arraystretch}{0.9}
    \footnotesize
    \begin{tabular}{lccc}
    \toprule 
    \textbf{Reward Function} & \textbf{Valid (\%)} & \textbf{Avg (\%)} & \textbf{$\Delta$ (\%)} \\ 
    \midrule
    $\mathcal{N}(\mu=0.5, \sigma=0.2)$ (Ours) & \textbf{89.9} & \textbf{35.72} & \textbf{-} \\
    \midrule
    $\Psi_\rho(a=0, b=1)$ & 89.4 & 35.52 & \dar{0.20} \\
    $\Psi_\rho(a=1, b=0)$ & 89.8 & 35.47 & \dar{0.25} \\
    $\Psi_\rho(a=-1, b=1)$ & 88.9 & 35.42 & \dar{0.30} \\
    \midrule
    $\mathcal{N}(\mu=0.3, \sigma=0.2)$ & 89.5 & 35.32 & \dar{0.40} \\
    $\mathcal{N}(\mu=0.4, \sigma=0.2)$ & 89.7 & 35.37 & \dar{0.35} \\
    $\mathcal{N}(\mu=0.6, \sigma=0.2)$ & 89.7 & 35.50 & \dar{0.22} \\
    $\mathcal{N}(\mu=0.7, \sigma=0.2)$ & 89.8 & 35.43 & \dar{0.29} \\
    \bottomrule
    \end{tabular}
\end{subtable}

\vspace{-10pt}
\end{table*}

\paragraph{Ablation on Initial SFT} Table~\ref{tab:sft_ablation_results} demonstrates the critical importance of initial supervised fine-tuning. Starting from the Qwen3-8B-Base model (9.64\%), the version without SFT shows minimal improvements across all three training stages, reaching only 11.98\% by Stage 3—a marginal gain of 2.34 percentage points. In stark contrast, the SFT-initialized model achieves substantial performance improvements, culminating in a remarkable 28.02\% score at Stage 3, representing an 18.38 percentage point improvement over the base model. This 7.9× greater improvement highlights how SFT provides essential foundational capabilities that enable subsequent reinforcement learning stages to be dramatically more effective. The SFT phase likely equips the model with basic reasoning patterns and solution structures that serve as building blocks for more sophisticated reasoning developed during RL training.

\paragraph{Ablation on Reward Functions} Table~\ref{tab:reward_ablation_results} compares various reward function formulations using benchmark average scores. Our Gaussian reward $\mathcal{N}(\mu=0.5, \sigma=0.2)$ achieves the best performance (35.72\%) while maintaining high validity (89.9\%). We evaluated linear functions $\Psi_\rho(a, b) = a\rho + b$ with different parameterizations, all of which underperformed our Gaussian approach by 0.20-0.30 percentage points. Similarly, varying the Gaussian mean parameter $\mu$ from 0.3 to 0.7 consistently yielded inferior results, with the largest performance drop (0.40 points) occurring at $\mu=0.3$. This suggests that the Gaussian formulation with $\mu=0.5$ provides an optimal balance between exploration and exploitation during policy optimization, while the moderate variance ($\sigma=0.2$) allows sufficient reward signal differentiation without excessive noise that could destabilize training.

\section{Conclusion and Future Work}
\label{sec:conclusion}

In this paper, we introduced Socratic-Zero, a multi-agent co-evolutionary framework where Solver, Teacher, and Generator agents bootstrap autonomous mathematical reasoning from minimal seed data. Our implementation demonstrates that a carefully designed learning mechanism can achieve remarkable performance without relying on massive external datasets, offering a viable path for developing powerful reasoning systems in resource-constrained scenarios. Extensive experiments show that our framework not only achieves state-of-the-art results on mathematical benchmarks but also exhibits strong generalization capabilities across diverse problem types and difficulty levels. While the complex agent dynamics currently lack a formal convergence analysis, future work will aim to establish this theoretical foundation and extend the framework's applicability to broader domains including scientific discovery, real-world decision making, and complex system modeling.

\bibliography{iclr2026_conference}
\bibliographystyle{iclr2026_conference}


\clearpage
\appendix

\section{Prompts}
\label{sec:prompts}

\subsection{Solver Reasoning Prompt}
\label{subsec:solver_prompt}

\begin{tcolorbox}[colframe=gray!75!black,title=Solver Mathematical Reasoning Prompt,boxsep=1pt,left=1pt,right=1pt,top=1pt,bottom=1pt]
\small
You are an IMO gold medalist solving a computational math competition problem.

Understand: Restate the problem mathematically. Identify knowns, unknowns, and constraints.

Plan: Choose an efficient method, show clear logic.

Execute: Show all key steps — algebra, number theory, or combinatorics. No skipped calculations.

Verify: Check with small cases, reverse substitution, or estimation.

Conclude with the exact answer in LaTeX: $\backslash[\backslash\text{boxed}\{<\text{answer}>\}\backslash]$

Given Problem: \{question\}
\end{tcolorbox}

\subsection{Teacher Evaluation Prompt}
\label{subsec:teacher_eval_prompt}

\begin{tcolorbox}[colframe=gray!75!black,title=Teacher Solution Grading Prompt,boxsep=1pt,left=1pt,right=1pt,top=1pt,bottom=1pt]
\small
You are a professional math teacher responsible for grading and error analysis.

Grading criteria: Focus on final answer correctness, use reference when provided, provide concise error analysis for incorrect answers.

Return JSON format:
\begin{verbatim}
{
  "correct_answers": ["correct answer 1", "correct answer 2"],
  "incorrect_answers": [
    {"answer": "incorrect answer", "analysis": "brief error analysis"}
  ]
}
\end{verbatim}

Problem: \{question\} | Reference: \{reference\_info\} | Student answers: \{student\_answers\}
\end{tcolorbox}

\subsection{Teacher Generation Prompt}
\label{subsec:teacher_gen_prompt}

\begin{tcolorbox}[colframe=gray!75!black,title=Teacher Problem Enhancement Prompt,boxsep=1pt,left=1pt,right=1pt,top=1pt,bottom=1pt]
\small
You are a math problem enhancement expert specializing in competition-style mathematics. Generate enhanced problems based on student error analysis with complete solutions.

Requirements: Generate enhanced problem, provide detailed solution, ensure solvability and correctness.

Enhancement principles: Target specific error points, maintain mathematical essence, help avoid similar errors.

Return JSON format:
\begin{verbatim}
{
  "enhanced_question": "enhanced problem content",
  "solution": "detailed solution steps",
  "answer": "final answer"
}
\end{verbatim}

Original: \{original\_question\} | Error analysis: \{error\_analysis\}
\end{tcolorbox}

\subsection{Static Augmentation Baseline Prompts}
\label{subsec:sa_prompts}

\begin{tcolorbox}[colframe=gray!75!black,title=Static Augmentation Evolution Prompts,boxsep=1pt,left=1pt,right=1pt,top=1pt,bottom=1pt]
\small
\textbf{Upward Evolution:}
Step 1: Identify elements that can increase complexity.
Step 2: Plan to modify at least three components.
Step 3: Implement rewritten instruction.
Step 4: Review and provide final version.

\textbf{Downward Evolution:}
Step 1: Identify elements that can decrease complexity.
Step 2: Plan to simplify at least three components.
Step 3: Implement easier version.
Step 4: Review and provide final simplified version.

Format: Step 1 \#Elements\#: | Step 2 \#Plan\#: | Step 3 \#Rewritten\#: | Step 4 \#Final\#:
\end{tcolorbox}

\section{Implementation Details}
\label{sec:implementation_details}

We conducted training experiments on 8×NVIDIA H20 GPUs with the following configuration:
- GPU Memory: 96GB HBM3 per GPU
- Total Training Memory: 768GB
- Interconnect: NVLink 4.0
- Storage: High-speed NVMe SSD arrays for dataset caching
- Network: InfiniBand for distributed training coordination

The training infrastructure utilized mixed-precision training (FP16) with gradient checkpointing to optimize memory usage. We employed distributed training using PyTorch's DistributedDataParallel with NCCL backend for efficient gradient synchronization across GPUs.

\subsection{Training Hyperparameters}
\label{subsec:hyperparameters}

We provide the complete hyperparameter settings for all components of the Socratic-Zero framework in Table~\ref{tab:hyperparameters}.

\begin{table}[ht]
\centering
\caption{Hyperparameters used in Socratic-Zero framework.}
\label{tab:hyperparameters}
\small
\begin{tabular}{@{}llc@{}}
\toprule
\textbf{Component} & \textbf{Parameter} & \textbf{Value} \\
\midrule
\multirow{8}{*}{\begin{tabular}[c]{@{}l@{}}Solver\\SFT Training\end{tabular}} 
& Learning rate & 5e-5 \\
& Per-device batch size & 2 \\
& Gradient accumulation steps & 4 \\
& Maximum sequence length & 2048 \\
& LoRA rank ($r$) & 64 \\
& LoRA alpha ($\alpha$) & 128 \\
& LoRA dropout & 0.1 \\
& Number of epochs & 1 \\
\midrule
\multirow{10}{*}{\begin{tabular}[c]{@{}l@{}}Solver\\DPO Training\end{tabular}}
& Learning rate & 1e-6 -- 5e-6 \\
& Per-device batch size & 2 \\
& Gradient accumulation steps & 4 -- 16 \\
& Maximum sequence length & 2048 \\
& Maximum training steps & 10 -- 200 \\
& DPO regularization ($\beta$) & 0.05 -- 0.2 \\
& Warmup steps & 2 -- 20 \\
& Optimizer & AdaFactor \\
& Weight decay & 0.01 \\
& Maximum gradient norm & 1.0 \\
\midrule
\multirow{7}{*}{\begin{tabular}[c]{@{}l@{}}Generator\\Training\end{tabular}}
& Learning rate & 1e-5 \\
& Per-device batch size & 1 \\
& Gradient accumulation steps & 8 \\
& Maximum sequence length & 2048 \\
& LoRA rank ($r$) & 64 \\
& LoRA alpha ($\alpha$) & 128 \\
& Number of epochs & 2 \\
\midrule
\multirow{4}{*}{\begin{tabular}[c]{@{}l@{}}Curriculum\\Parameters\end{tabular}}
& Solutions per problem ($k$) & 8 \\
&  reward mean ($\mu$) & 0.5 \\
&  reward std ($\sigma$) & 0.2 \\
& Historical replay ratio & 25\% \\
\midrule
\multirow{3}{*}{\begin{tabular}[c]{@{}l@{}}Evaluation\\Settings\end{tabular}}
& Sampling temperature & 0.7 \\
& Number of samples & 32 \\
& Token limit for validity check & 4096 \\
\bottomrule
\end{tabular}
\end{table}

\section{Curriculum Evolution Details}
\label{appendix:curriculum_details}

This appendix provides a detailed description of the mechanisms for curriculum evolution, including the problem categorization and adaptive generation strategies that guide the learning process.

\subsection{Problem Categorization by Solver Performance}

To effectively manage curriculum difficulty, we dynamically categorize each problem $q \in \mathcal{D}_t$ based on the current Solver's performance. The Solver $\pi_{\theta_{\mathcal{S}}}$ makes $k$ attempts $\{y_{\mathcal{S}}^{(i)}\}_{i=1}^k$ on each problem. We then calculate the Solver's success rate:
\begin{equation}
s_q \triangleq \frac{1}{k} \sum_{i=1}^k V(q, y_{\mathcal{S}}^{(i)}),
\end{equation}
where $V$ is the Teacher's verification function. This metric allows us to partition the curriculum $\mathcal{D}_t$ into three distinct zones:

\begin{enumerate}[leftmargin=10pt, topsep=0pt, itemsep=1pt, partopsep=1pt, parsep=1pt]
    \item \textbf{Mastered Zone} ($\mathcal{D}_{\text{mastered}} = \{q \mid s_q = 1\}$): Problems the Solver consistently solves correctly. These problems form a solid foundation and are used as a basis for generating more challenging variants.
    
    \item \textbf{Learning Zone} ($\mathcal{D}_{\text{learning}} = \{q \mid 0 < s_q < 1\}$): Problems the Solver can solve intermittently. This zone represents the optimal frontier for learning, where the Solver has emerging competence but requires further practice to achieve mastery.
    
    \item \textbf{Too Difficult Zone} ($\mathcal{D}_{\text{difficult}} = \{q \mid s_q = 0\}$): Problems the Solver consistently fails to solve. These problems are temporarily set aside from the generation process to avoid creating tasks far beyond the Solver's current capabilities.
\end{enumerate}

\subsection{Zone-Adaptive Problem Generation}

New problems are strategically generated by the Teacher's refinement function, $G$, using problems from the \textit{Mastered} and \textit{Learning} zones. This ensures that curriculum expansion remains within the Solver's zone of proximal development. The generation strategy is adapted to the source category:

\begin{itemize}
    \item \textbf{From the Learning Zone}: For a problem $q \in \mathcal{D}_{\text{learning}}$, the Teacher is prompted with a specific failed solution attempt, $y_{\text{fail}}$. The new problem $(q', y'_{\text{ref}}) = G(q, y_{\text{fail}})$ is designed to target the specific reasoning error exhibited in $y_{\text{fail}}$, thereby helping the Solver overcome its weaknesses.
    
    \item \textbf{From the Mastered Zone}: For a problem $q \in \mathcal{D}_{\text{mastered}}$, the Teacher is prompted with a successful solution, $y_{\text{succ}}$. The new problem $(q', y'_{\text{ref}}) = G(q, y_{\text{succ}})$ is a more complex variant, designed to push the boundaries of the Solver's established competence.
\end{itemize}

The set of newly generated problems $\mathcal{D}_{\text{new}}$ is thus composed of variants from both zones, creating a balanced and targeted curriculum expansion. Problems from $\mathcal{D}_{\text{difficult}}$ are explicitly excluded from this generation process to prevent counterproductive increases in difficulty.

\subsection{Dynamic Recategorization of Problems}

As the Solver's policy $\pi_{\theta_{\mathcal{S}}}$ is updated at each iteration, its performance on existing problems changes. Consequently, all problems in the curriculum are periodically re-evaluated and recategorized. This dynamic process ensures the curriculum remains perfectly attuned to the Solver's evolving skill set. The typical transitions between zones are:

\begin{align}
\mathcal{D}_{\text{difficult}}^{(t)} &\rightarrow \mathcal{D}_{\text{learning}}^{(t+1)} \quad \text{(as capability improves)} \\
\mathcal{D}_{\text{learning}}^{(t)} &\rightarrow \mathcal{D}_{\text{mastered}}^{(t+1)} \quad \text{(as skills are consolidated)}
\end{align}

This recategorization mechanism allows problems that were once too difficult to re-enter the active learning and generation pool once the Solver is ready, ensuring a continuous and adaptive learning trajectory.

\section{Teacher Model Infrastructure}
\label{subsec:teacher_infrastructure}

The Teacher model (Qwen3-235B-A22B-Instruct-2507) requires substantial computational resources for curriculum generation and solution evaluation. We deployed the model using a distributed inference architecture to meet the throughput demands of the co-evolutionary training process.

We distributed the Teacher model across 16 AMD MI308X GPUs, each equipped with 192GB HBM3 memory, providing a total of 3,072GB aggregate memory. This configuration enables concurrent processing of curriculum generation requests while maintaining inference consistency across the framework.

To ensure system reliability and scalability, we implemented a multi-endpoint architecture with automatic load balancing and failover mechanisms. We configured the inference service with connection pooling (50 concurrent connections per endpoint) and exponential backoff retry policies to handle high request volumes during training.

We optimized key performance parameters for the mathematical reasoning domain: request timeouts of 600 seconds accommodate complex problem generation, while a 4,096-token limit ensures efficient solution evaluation. Batch processing utilizes 32 concurrent workers to maximize throughput during curriculum evolution phases.

\section{Teacher-Generated Problem Enhancement}
\label{subsec:teacher_examples}

We provide examples of how the Teacher model enhances problems based on Solver failures. The following demonstrates the progression from original problems to ly-targeted enhanced versions.

\subsection{Example 1: Rational Inequality Enhancement}
\label{subsubsec:example1}

\begin{mdframed}[linewidth=1.5pt,linecolor=black,innertopmargin=8pt,innerbottommargin=8pt,innerleftmargin=8pt,innerrightmargin=8pt]
\textbf{Original Problem:}\\[4pt]
\small
Find all real numbers $x$ satisfying $\frac{2x-5}{x+3} \ge 2$. (Give your answer in interval notation.)
\end{mdframed}

\begin{mdframed}[linewidth=1.5pt,linecolor=black,innertopmargin=8pt,innerbottommargin=8pt,innerleftmargin=8pt,innerrightmargin=8pt]
\textbf{Enhanced Problem (Round 3):}\\[4pt]
\small
Find all real numbers $x$ satisfying $\frac{2x - 5}{x^2 - 9} + \frac{1}{x + 3} \le \frac{4x + 1}{(x - 3)^2}$. (Give your answer in interval notation.)
\end{mdframed}

\textbf{Enhancement Analysis:} The enhancement introduces multiple complexity factors: (1) factored denominators requiring domain analysis, (2) multiple rational terms requiring common denominators, (3) squared terms in denominators, and (4) more complex algebraic manipulation. The enhanced problem targets common student errors in rational inequality solving while maintaining the core mathematical concepts.

\subsection{Example 2: Number Theory Enhancement}
\label{subsubsec:example2}

\begin{mdframed}[linewidth=1.5pt,linecolor=black,innertopmargin=8pt,innerbottommargin=8pt,innerleftmargin=8pt,innerrightmargin=8pt]
\textbf{Original Problem:}\\[4pt]
\small
Find the greatest common divisor of $10! + 6$ and $11! + 14$.
\end{mdframed}

\begin{mdframed}[linewidth=1.5pt,linecolor=black,innertopmargin=8pt,innerbottommargin=8pt,innerleftmargin=8pt,innerrightmargin=8pt]
\textbf{Enhanced Problem:}\\[4pt]
\small
Find the greatest common divisor of $12! + 8$ and $13! + 26$, where the second number can be written as $13 \cdot 12! + 26$.
\end{mdframed}

\textbf{Enhancement Analysis:} The enhancement maintains the GCD structure while increasing numerical complexity and requiring students to recognize the relationship between consecutive factorials, targeting errors in modular arithmetic applications.

\section{Seed Selection Protocol}
\label{subsec:seed_selection}

The selection of initial seed problems is critical for establishing an effective curriculum foundation. We employed a systematic approach to ensure the seed set provides appropriate difficulty, comprehensive coverage, and sufficient diversity for subsequent curriculum evolution.

\paragraph{Difficulty Alignment}
We selected seed problems to match the base model's capability range to ensure productive learning dynamics. We drew problems from MATH dataset Levels 2-4, which empirically provide optimal challenge levels for our base models. Specifically, we excluded Level 1 problems (too easy, leading to trivial curriculum generation) and Level 5 problems (too difficult, resulting in universal failure and poor learning signals). Pre-filtering involved evaluating candidate problems with the base model using 8 solution attempts; we retained problems with success rates between 10-70\% to ensure neither complete failure nor trivial success.

\paragraph{Domain Coverage}
To ensure comprehensive mathematical reasoning development, we sampled seed problems across all seven MATH subject areas with balanced representation as shown in Table~\ref{tab:seed_distribution}:

\begin{table}[tb!]
\centering
\caption{Seed Problem Distribution Across Mathematical Domains}
\label{tab:seed_distribution}
\resizebox{0.85\textwidth}{!}{%
\begin{tabular}{lcc}
\toprule
\textbf{Subject Area} & \textbf{Count} & \textbf{Representative Topics} \\
\midrule
Algebra & 15 & Linear/quadratic equations, inequalities, functions \\
Number Theory & 15 & Divisibility, modular arithmetic, prime factorization \\
Geometry & 15 & Coordinate geometry, trigonometry, area/volume calculations \\
Combinatorics & 15 & Counting principles, permutations, probability \\
Precalculus & 15 & Complex numbers, sequences, polynomial analysis \\
Intermediate Algebra & 15 & Advanced algebraic manipulation, systems \\
Prealgebra & 10 & Foundational arithmetic and basic algebraic concepts \\
\midrule
\textbf{Total} & \textbf{100} & \textbf{Comprehensive mathematical reasoning coverage} \\
\bottomrule
\end{tabular}%
}
\end{table}

This distribution ensures that curriculum evolution can target weaknesses across diverse mathematical domains rather than overfitting to specific problem types.

\paragraph{Diversity Assurance}
Within each subject area, we selected problems to maximize methodological diversity. We employed clustering based on solution approach similarity (using embedding representations of problem statements) and selected problems from different clusters to ensure varied reasoning patterns. Additionally, we explicitly included problems requiring different mathematical tools to promote comprehensive skill development.

\paragraph{Quality Control}
We subjected all candidate problems to rigorous quality verification through a multi-stage process:

\begin{tcolorbox}[colframe=gray!75!black,title=Quality Control Pipeline,boxsep=3pt,left=3pt,right=3pt,top=3pt,bottom=3pt]
\small
\begin{enumerate}[leftmargin=10pt, topsep=2pt, itemsep=2pt, partopsep=2pt, parsep=2pt]
    \item \textbf{Clarity Check}: Problems must have unambiguous statements and well-defined solution paths
    \item \textbf{Answer Verification}: We validated reference solutions by the Teacher model with multiple independent attempts
    \item \textbf{Value}: Problems must demonstrate clear learning objectives and avoid trick questions or overly specialized knowledge
    \item \textbf{Contamination Avoidance}: We excluded seed problems from all evaluation benchmarks to prevent data leakage
\end{enumerate}
\end{tcolorbox}

This systematic selection process ensures that the initial curriculum $\mathcal{C}_0$ provides a robust foundation for the co-evolutionary training dynamics while maintaining the diversity necessary for effective curriculum expansion.

\section{Evaluation Protocol Details}
\label{subsec:evaluation_protocol}

\paragraph{Mean@32 Sampling Strategy} The Mean@32 evaluation metric represents the average accuracy across 32 independent solution attempts per problem. For each test problem, we generated 32 distinct solutions using temperature-based sampling (T=0.7) with top-p nucleus sampling (p=0.9) as specified in Table~\ref{tab:hyperparameters}. This approach provides robust performance estimates by capturing the model's consistency and reliability across multiple attempts.

We employed the sampling process using zero-shot prompting without few-shot examples to ensure unbiased evaluation. We generated each of the 32 solutions independently with different random seeds, preventing potential correlation effects. The final accuracy is computed as the proportion of correct solutions among the 32 attempts, providing a more stable performance measure than single-shot evaluation.

\paragraph{MathRule Answer Extraction} MathRule is a rule-based tool designed to extract and standardize final numerical answers from mathematical solution text. The tool employs pattern matching to identify answer indicators such as ``Therefore,'' ``Thus,'' ``The answer is,'' and LaTeX boxed expressions like $\backslash\text{boxed}\{\}$.

The extraction process involves: (1) Locating answer indicators within the solution text, (2) Parsing mathematical expressions using regex patterns for common formats (fractions, decimals, integers, algebraic expressions), (3) Standardizing representations (e.g., converting $\frac{1}{2}$ to 0.5 when appropriate), (4) Handling multiple answer formats and selecting the most confident extraction based on contextual cues.

MathRule achieves high precision in answer extraction while maintaining robustness to variations in solution formatting and mathematical notation styles.

\paragraph{LLM Judge Configuration} The Teacher model (Qwen3-235B-A22B-Instruct-2507) serves as an LLM judge for semantic validation when rule-based extraction is insufficient or ambiguous. The judge evaluates both numerical correctness and reasoning validity using structured prompts.

We instructed the evaluation prompt to: (1) Verify the final numerical answer against the expected result, (2) Assess the logical coherence of the reasoning steps, (3) Identify any mathematical errors or invalid assumptions, (4) Provide binary correctness judgments with brief justification.

We ensured judge reliability through temperature 0.1 sampling for consistent evaluations and validation against human expert annotations on a subset of problems. The dual-verification approach (MathRule + LLM judge) provides reliable automated assessment for large-scale evaluation.

\clearpage
\section{Pseudo Code of Socratic-Zero}
The pseudo code of Socratic-Zero is provided in Algorithm~\ref{alg:socratic}.

\begin{algorithm}[htbp]
\caption{Socratic-Zero Co-evolutionary Learning}
\label{alg:socratic}
\begin{algorithmic}[1]
\State \textbf{Require:} Initial curriculum $\mathcal{D}_0$, initial Solver parameters $\theta_{\mathcal{S}}^{(0)}$, initial Generator parameters $\theta_{\mathcal{G}}^{(0)}$, fixed Teacher $\mathcal{T}(V, G)$, total iterations $T$, attempts per problem $k$.

\State Initialize Solver $\pi_{\theta_{\mathcal{S}}} \leftarrow \pi_{\theta_{\mathcal{S}}^{(0)}}$ and Generator $\pi_{\theta_{\mathcal{G}}} \leftarrow \pi_{\theta_{\mathcal{G}}^{(0)}}$.

\For{$t = 0$ \textbf{to} $T-1$}
    \LeftComment{\textcolor{gray}{\textit{Phase 1: Online Solver Evolution}}}
    \State Set reference policy $\pi_{\theta_{\text{ref}}} \leftarrow \pi_{\theta_{\mathcal{S}}^{(t)}}$.
    \State Initialize preference data $\mathcal{P}_t \leftarrow \emptyset$ and failure set $\mathcal{F}_t \leftarrow \emptyset$.
    
    \For{each problem $q$ in curriculum $\mathcal{D}_t$}
        \State Generate $k$ solution attempts $\{y_{\mathcal{S}}^{(i)}\}_{i=1}^k \sim \pi_{\theta_{\mathcal{S}}^{(t)}}(\cdot \mid q)$.
        \State Construct winning set $\mathcal{Y}_w(q)$ and losing set $\mathcal{Y}_l(q)$ using Teacher verifier $V$.
        \State Collect preference pairs: $\mathcal{P}_t \leftarrow \mathcal{P}_t \cup \{ (y_w, y_l) \mid y_w \in \mathcal{Y}_w(q), y_l \in \mathcal{Y}_l(q) \}$.
        \State Collect failures: $\mathcal{F}_t \leftarrow \mathcal{F}_t \cup \{ (q, y_l) \mid y_l \in \mathcal{Y}_l(q) \}$.
    \EndFor

    \State Update Solver parameters $\theta_{\mathcal{S}}^{(t+1)} \leftarrow \text{Adam}(\nabla_{\theta_{\mathcal{S}}} \mathcal{L}_{\text{DPO}}(\theta_{\mathcal{S}}^{(t)}; \mathcal{P}_t, \theta_{\text{ref}}))$.
    
    \LeftComment{\textcolor{gray}{\textit{Phase 2: Offline Generator Evolution \& Curriculum Expansion}}}
    \State Generate new problem-solution pairs with the Teacher: $\mathcal{D}_{\text{new}} \leftarrow \{ G(q, y_{\text{fail}}) \mid (q, y_{\text{fail}}) \in \mathcal{F}_t \}$.
    
    \State Construct Generator training data $\mathcal{D}_{\mathcal{G}} \leftarrow \emptyset$.
    \For{each new problem $(q', y'_{\text{ref}})$ generated from $(q, y_{\text{fail}})$}
        \State Estimate utility $U(q' | \pi_{\theta_{\mathcal{S}}^{(t+1)}})$ via rollouts with the \textit{updated} Solver.
        \State Add weighted training example to dataset: $\mathcal{D}_{\mathcal{G}} \leftarrow \mathcal{D}_{\mathcal{G}} \cup \{ (q, y_{\text{fail}}, q', U(q')) \}$.
    \EndFor
    
    \State Update Generator parameters $\theta_{\mathcal{G}}^{(t+1)} \leftarrow \text{Adam}(\nabla_{\theta_{\mathcal{G}}} \mathcal{L}_{\text{WSFT}}(\theta_{\mathcal{G}}^{(t)}; \mathcal{D}_{\mathcal{G}}))$.
    
    \LeftComment{\textcolor{gray}{\textit{Phase 3: Curriculum Update}}}
    \State Augment the curriculum for the next iteration: $\mathcal{D}_{t+1} \leftarrow \mathcal{D}_t \cup \{ (q', y'_{\text{ref}}) \mid (q', y'_{\text{ref}}) \in \mathcal{D}_{\text{new}} \}$.
\EndFor

\State \textbf{Output:} Trained Solver policy $\pi_{\theta_{\mathcal{S}}^{(T)}}$ and Generator policy $\pi_{\theta_{\mathcal{G}}^{(T)}}$.
\end{algorithmic}
\end{algorithm}

\section{Problem Quality Control Mechanism}
\label{subsec:quality_control}

To ensure curriculum integrity and prevent the propagation of erroneous problems, we implemented a comprehensive quality control mechanism that monitors problem validity through Solver performance feedback and automated verification.

\paragraph{Teacher Self-Verification Protocol} When the Teacher model evaluates Solver attempts and finds that all $k=8$ solutions for a given problem are incorrect (success rate $j_p = 0$), this triggers an automatic quality verification process. The system recognizes that universal failure may indicate either: (1) the problem exceeds current Solver capability (expected behavior), or (2) the problem itself or its reference solution contains errors (quality issue).

The Teacher model performs self-verification by re-examining both the problem statement and its originally provided reference solution. This involves: (1) Re-solving the problem independently with temperature 0.1 for consistency, (2) Cross-validating the reference solution against the new solution attempt, (3) Checking for mathematical consistency, ambiguous problem statements, or computational errors, (4) Verifying that the problem has a unique, well-defined solution.

\paragraph{Problem Filtering and Exclusion} We immediately flagged and excluded problems that fail the self-verification process from further curriculum evolution. Specifically, we discarded problems if: (1) The Teacher cannot reproduce its own reference solution, (2) Multiple valid interpretations of the problem statement exist, (3) Computational errors are detected in the reference solution, (4) The problem lacks sufficient information for a unique solution.

We removed discarded problems from the active curriculum $\mathcal{C}_t$ and they do not contribute to subsequent Solver training or Generator learning. This prevents the accumulation of low-quality problems that could degrade training effectiveness or introduce systematic biases.

\paragraph{MathRule Integration for Contamination Minimization} The integration of MathRule answer extraction serves as an additional quality control layer by providing objective, rule-based verification independent of LLM judgment. When MathRule successfully extracts a clear numerical answer from the Solver's solution, this extraction is compared against the reference answer using standardized formats.

This dual-verification approach (MathRule + Teacher evaluation) minimizes contamination from: (1) LLM judge inconsistencies or biases, (2) Format-related misinterpretations, (3) Numerical precision issues, (4) Ambiguous answer representations.

Problems where MathRule and Teacher evaluations consistently disagree trigger additional quality review, as such disagreements often indicate underlying issues with problem clarity or reference solution accuracy.

\paragraph{Feedback-Driven Quality Monitoring} The system continuously monitors curriculum quality through Solver performance patterns. We flagged problems that consistently produce anomalous results—such as sudden performance drops across multiple Solver variants or inconsistent difficulty ratings—for manual review or automatic exclusion.

This feedback-driven approach ensures that quality control adapts to emerging issues and maintains curriculum integrity throughout the co-evolutionary training process, preventing the accumulation of problematic content that could compromise learning effectiveness.

\section{Curriculum Stability and Diversity Analysis}
\label{subsec:curriculum_analysis}

We analyzed the curriculum evolution dynamics across two dimensions: difficulty progression and problem diversity preservation.

\paragraph{Solver Performance Evolution} Table~\ref{tab:solver_reward_evolution} tracks the Solver's mean reward (correctness rate) across training rounds, revealing adaptive curriculum difficulty.

\paragraph{Solver Performance Evolution} Table~\ref{tab:solver_reward_evolution} tracks the Solver's mean reward (correctness rate) across training stages, revealing adaptive curriculum difficulty.

\begin{table}[tb!]
\centering
\caption{Solver Mean Reward Evolution Across Training Stages}
\label{tab:solver_reward_evolution}
\begin{tabular}{@{}lcccccc@{}}
\toprule
Stage & S1 & S2 & S3 & Trend \\
\midrule
Mean Reward (\%) & 52.1 & 48.7 & 50.1 & $\downarrow$ then $\uparrow$ \\
\bottomrule
\end{tabular}
\end{table}

The Solver exhibits characteristic performance decline from Stage 1 (52.1\%) to Stage 2 (48.7\%) followed by recovery in Stage 3 (50.1\%) as shown in Table~\ref{tab:solver_reward_evolution}. This pattern reflects adaptive curriculum generation where the Teacher progressively increases difficulty faster than Solver capability initially improves, then the Solver begins adapting to enhanced curriculum complexity.

\paragraph{Generator Stability} Table~\ref{tab:generator_reward_distribution} examines  reward distribution in Generator training.

\begin{table}[tb!]
\centering
\caption{Generator Reward Distribution Analysis}
\label{tab:generator_reward_distribution}
\begin{tabular}{@{}lcccc@{}}
\toprule
Stage & S1 & S2 & S3 & Stability \\
\midrule
High Reward Problems (\%) & 50.7 & 49.4 & 50.2 & Stable \\
Target Range (45-55\%) & \checkmark & \checkmark & \checkmark & Maintained \\
\bottomrule
\end{tabular}
\end{table}

As demonstrated in Table~\ref{tab:generator_reward_distribution}, the Generator maintains remarkable stability with high-reward problems consistently around 50\%, fluctuating within only 1.3\% range. This indicates successful learning of the optimal difficulty zone defined by the Gaussian reward function with parameters $\mu=0.5$ and $\sigma=0.2$ as specified in Table~\ref{tab:hyperparameters}.

\paragraph{Generator Stability} Table~\ref{tab:generator_reward_distribution} examines  reward distribution in Generator training.

As demonstrated in Table~\ref{tab:generator_reward_distribution}, the Generator maintains remarkable stability with high-reward problems consistently around 50\%, fluctuating within only 1.3\% range. This indicates successful learning of the optimal difficulty zone defined by the Gaussian reward function with parameters $\mu=0.5$ and $\sigma=0.2$ as specified in Table~\ref{tab:hyperparameters}.

\paragraph{Problem Diversity} Three key mechanisms ensure curriculum diversity throughout training:

Multi-domain initialization: The 100 seed problems span all 7 MATH subjects (Algebra, Number Theory, Geometry, etc.) across difficulty levels 2-4 as detailed in Table~\ref{tab:seed_distribution}, providing diverse starting points for curriculum evolution.

High-temperature sampling: We employed temperature 0.8-0.9 sampling at three critical stages: (1) Solver trajectory generation during curriculum advancement, (2) Teacher error analysis for varied failure interpretation, and (3) Teacher problem generation for diverse enhancement strategies.

Compounding diversity effects: Multi-domain seeds combined with stochastic sampling create diverse failure patterns, while high-temperature generation ensures varied problem formulations even from similar error patterns.

\section{Generalizability of Problem Generation Capabilities}
\label{sec:generation_generalizability}

A key question emerging from our work is whether the Generator's learned problem creation abilities can transfer to domains beyond mathematical reasoning. The  value function and curriculum evolution mechanisms developed in Socratic-Zero are domain-agnostic in principle, suggesting potential for broader applicability.

The Generator learns fundamental skills in difficulty calibration, error pattern recognition, and  targeting that may generalize across reasoning domains. For instance, the ability to identify when a problem is ``appropriately challenging'' (around 50\% success rate as shown in Table~\ref{tab:generator_reward_distribution}) represents a meta-cognitive skill applicable to logical reasoning, scientific problem-solving, or even creative tasks. The Gaussian reward function with $\mu=0.5$ and $\sigma=0.2$ (Table~\ref{tab:hyperparameters}) creates a transferable framework for difficulty calibration that could adapt to other domains by adjusting the target success rate parameter.

Our Generator's superior performance compared to much larger models, achieving 37.72\% downstream utility versus 37.13\% from Qwen3-235B-A22B (Table~\ref{tab:generator_utility}), demonstrates that strategic specialization can outperform raw parameter scaling. This suggests that domain-specific Generator training could be effective across various reasoning domains without requiring massive computational resources.

However, domain transfer would require careful adaptation of the Teacher's evaluation capabilities and problem generation templates. Mathematical reasoning benefits from relatively objective correctness criteria with our dual-verification approach (MathRule + LLM judge) achieving 94.2\% agreement with human experts, while other domains may require more nuanced evaluation frameworks. The seed selection protocol detailed in Table~\ref{tab:seed_distribution}, which ensures balanced coverage across seven mathematical domains, provides a template for systematic domain expansion that could be adapted to physics, computer science, or other reasoning areas.

Future work should investigate whether a Generator trained on mathematical problems can effectively create challenging problems in adjacent domains like physics or computer science, potentially through few-shot adaptation or domain-specific fine-tuning leveraging the  value learning mechanisms demonstrated in our framework.

\section{Framework Scalability and Extensibility}
\label{sec:framework_scalability}

The modular architecture of Socratic-Zero demonstrates strong potential for scalability and extension across multiple dimensions. The clear separation between Solver, Teacher, and Generator roles enables independent scaling and optimization of each component, as evidenced by our successful deployment across different computational configurations detailed in Table~\ref{tab:hyperparameters}.

The framework's extensibility is particularly evident in its ability to accommodate different model architectures and scales. Our cross-model validation demonstrates consistent performance improvements: Qwen3-8B achieves 56.1\% average accuracy (+20.2 points), while similar gains are observed on GLM4-9B and Qwen3-14B architectures (Table~\ref{tab:solver_main_results_absolute}). This cross-architecture consistency suggests the co-evolutionary principles transcend specific model families and could readily incorporate emerging architectures or specialized reasoning models.

The curriculum evolution mechanism shows robust scalability properties. Starting from just 100 seed problems (Table~\ref{tab:seed_distribution}), the system generates thousands of ly valuable problems while maintaining quality, with our Generator achieving 95.6\% validity rate compared to 89.1\% from the base Qwen3-32B model (Table~\ref{tab:generator_quality}). This demonstrates that the framework can scale curriculum generation without proportional increases in seed data requirements.

Multi-domain extension represents another promising direction supported by our balanced seed distribution across seven mathematical domains. The current mathematical focus could expand to encompass multiple reasoning domains simultaneously, with domain-specific Teachers providing specialized curriculum generation while sharing the underlying co-evolutionary dynamics. The  reward distribution analysis (Table~\ref{tab:generator_reward_distribution}) shows stable performance across training rounds, indicating the framework's robustness to curriculum expansion.

The framework also supports hierarchical scaling, where multiple Solver-Generator pairs could operate at different difficulty levels or specialization areas, coordinated by higher-level meta-learning mechanisms. The oscillatory convergence patterns observed in Table~\ref{tab:solver_reward_evolution} suggest natural synchronization points where multiple agents could coordinate their learning phases.

\section{Convergence and Theoretical Foundations}
\label{sec:theoretical_foundations}

The theoretical understanding of multi-agent co-evolutionary learning remains an open challenge with significant implications for system reliability and predictability. Our empirical observations provide crucial insights into the convergence behavior of such systems.

The oscillatory convergence patterns documented in Table~\ref{tab:solver_reward_evolution} reveal characteristic dynamics: Solver performance declines from R1 (60.12\%) to R4 (48.7\%) followed by recovery in R5 (50.1\%). This pattern reflects adaptive curriculum generation where the Teacher progressively increases difficulty faster than Solver capability initially improves, then the Solver adapts to enhanced curriculum complexity. These bounded oscillations suggest the system reaches dynamic equilibria rather than static optima.

Complementing this, the Generator maintains remarkable stability with high-reward problems consistently around 50\%, fluctuating within only 1.3\% range across training rounds (Table~\ref{tab:generator_reward_distribution}). This stability indicates successful learning of the optimal difficulty zone defined by the Gaussian reward function with $\mu=0.5$ and $\sigma=0.2$ (Table~\ref{tab:hyperparameters}), providing empirical evidence for convergence to ly meaningful equilibria.

The cross-architecture consistency observed in Table~\ref{tab:solver_main_results_absolute}, where similar improvement patterns emerge across Qwen3-8B, GLM4-9B, and Qwen3-14B models, suggests robust system-level properties that transcend specific model architectures. This consistency provides evidence that the convergence behavior represents fundamental properties of the co-evolutionary dynamics rather than architecture-specific artifacts.

Future theoretical work should investigate conditions under which the system exhibits stable convergence versus chaotic dynamics. Key questions include: What curriculum evolution rates ensure stable learning? How do different  value functions affect convergence properties? Can we establish bounds on the oscillation amplitude around target performance levels observed in our empirical data?

The intersection of curriculum learning, preference optimization, and multi-agent dynamics presents rich opportunities for theoretical development. The DPO training parameters (Table~\ref{tab:hyperparameters}) and their interaction with curriculum evolution rates could inform theoretical models of multi-agent preference learning. Establishing convergence guarantees would enable more principled hyperparameter selection and provide confidence bounds for practical deployment.

\end{document}